\documentclass[letterpaper, 10 pt, conference]{ieeetran}  
\IEEEoverridecommandlockouts                              

\usepackage{graphics}
\usepackage{float} 
\usepackage{caption}
\usepackage{subfigure}
\usepackage{epsfig}
\usepackage{mathptmx}
\usepackage{times}
\usepackage{amsmath} 
\usepackage{amssymb}  
\usepackage{array}
\usepackage{url} 
\usepackage{color}
\usepackage{cite} 
\usepackage[ruled,linesnumbered]{algorithm2e}  
\usepackage{siunitx}
\usepackage{alphalph}
\usepackage{tikz}  
\usetikzlibrary{tikzmark}
\usepackage{xcolor} 
\SetNlSty{bfseries}{\color{black}}{} 

\title{\LARGE \bf
	Graph Gain: A Concave-Hull Based Volumetric Gain \\ for Robotic Exploration
}

\author{
	Zezhou Sun,
    Huajun Liu, 
	Chengzhong Xu,
	Hui Kong
	\thanks{Zezhou Sun and Huajun Liu are with the School of Computer Science and Engineering, Nanjing University of Science and Technology, Nanjing 210094, Jiangsu, China (e-mail: sun\_zezhou@njust.edu.cn; liuhj@njust.edu.cn).}
	\thanks{Chengzhong Xu is with the State Key Laboratory of Internet of Things for Smart City (SKL-IOTSC), Department of Computer Science, University of Macau, Macau 999078, China (e-mail: czxu@um.edu.mo).}
	\thanks{Hui Kong is with The State Key Laboratory of Internet of Things for Smart City (SKL-IOTSC), Department of Electromechanical Engineering (EME), University of Macau, Macau 999078, China (e-mail: huikong@um.edu.mo).}
	\thanks{
		This work is supported by National Key Research and Development Program of China (No. 2019YFB2102100), the Science and Technology Development Fund of Macau SAR (File no. 0015/2019/AKP), Guangdong-Hong Kong-Macao Joint Laboratory of Human-Machine Intelligence-Synergy Systems (No. 2019B121205007), and the National Natural Science Foundation of China (No.61803083). }
}

\begin{document}
	
	\maketitle
	\thispagestyle{empty}
	\pagestyle{empty}
	
	\begin{abstract} 
	The existing volumetric gain for robotic exploration is calculated in the 3D occupancy map, while the sampling-based exploration method is extended in the reachable (free) space. 
	The inconsistency between them makes the existing calculation of volumetric gain inappropriate for a complete exploration of the environment.  
	To address this issue, we propose a concave-hull based volumetric gain in a sampling-based exploration framework. The concave hull is constructed based on the 
    viewpoints generated by Rapidly-exploring Random Tree (RRT) and the nodes that fail to expand. All space outside this concave hull is considered unknown. 
    The volumetric gain is calculated based on the viewpoints configuration rather than using the occupancy map. 
    With the new volumetric gain, robots can avoid inefficient or even erroneous exploration behavior caused by the inappropriateness of existing volumetric gain calculation methods. 
    Our exploration method is evaluated against the existing state-of-the-art RRT-based method in a benchmark environment \cite{cmuexploration}. 
    In the evaluated environment, the average running time of our method is about $38.4\%$ of the existing state-of-the-art method and our method is more robust. 
	\end{abstract}
	
	
	\section{Introduction}
    In an autonomous robotic exploration framework, mobile robots can navigate autonomously and build a complete pure 3D or 3D semantic model in an unknown environment. This involves an interleaved exploration module and a SLAM module. In the exploration stage, the robot can explore a set of destination candidates for future visits from the current location and rank them based on a predefined exploration gain, and then navigate to the destination with the highest rank. In the SLAM module, the robot can simultaneously do a mapping of the environment and localize itself in it, during which the environment model is updated with new observations.

\begin{figure}
		\centering
		\includegraphics[width=0.9\linewidth]{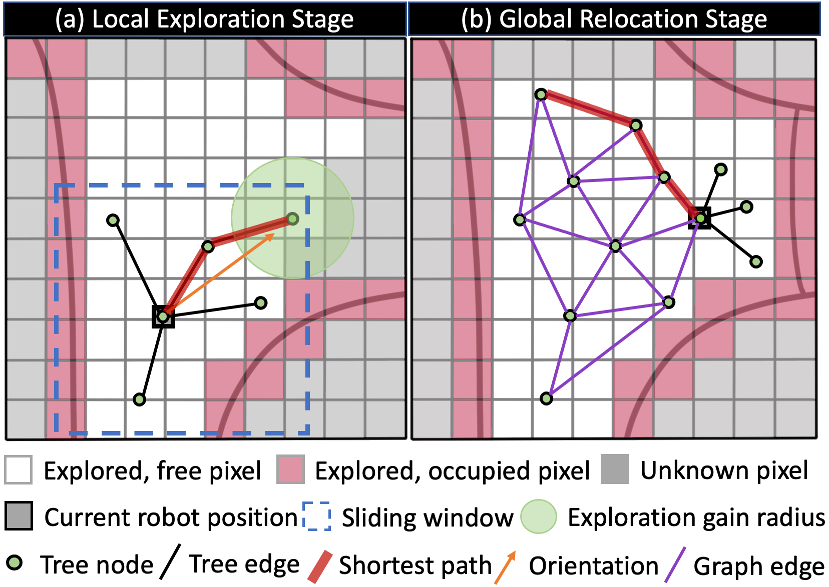}
		\caption{
		Illustration of the local exploration stage and the global relocation stage of the autonomous exploration framework.
        }
		\label{two-stage}
	\end{figure}

	
	The focus of this work is on exploration strategy and the mapping part can be implemented by a traditional SLAM method. Generally, the exploration module can be divided into a local exploration stage and a global relocation stage.
	In the local exploration stage, the local planner usually uses an exploration tree, e.g., a Rapidly-exploring Random Tree (RRT) \cite{lavalle1998rapidly}, to expand from the current robot location to the surrounding area. For efficiency, it usually finds tree nodes within a sliding window to limit the expansion range of the RRT. 
    The nodes of the RRT are regarded as the viewpoints sampled in the environment. 
    The volume of the unknown area around each viewpoint can be regarded as its volumetric gain. 
    The cumulative RRT branch length from the root to the viewpoint can be regarded as the distance to this viewpoint. The direction of the branch where the viewpoint is located can be regarded as the direction of exploration, as shown in Fig. \ref{two-stage}(a). 
    The volumetric gain, distance, and exploration direction are utilized to calculate the exploration gain.
    Robots tend to go to the viewpoint with the highest exploration gain and leave other unvisited candidate viewpoints for subsequent exploration.
    
    Simultaneously, all tree nodes are connected to form a Rapidly-exploring Random Graph (RRG) \cite{karaman2011sampling}, as shown in Fig.\ref{two-stage}(b). 
    When no more viewpoints with sufficiently high exploration gain are extracted within 2-3 consecutive sliding windows, the robot considers the local region as fully explored and then merges the local RRG into the global RRG. 
    It then switches to the global relocation stage to plan routes to the unvisited candidate viewpoints in the global RRG, shown as the red curve in Fig.\ref{two-stage}(b). 
    
    
\begin{figure}
		\centering
		\includegraphics[width=0.9\linewidth]{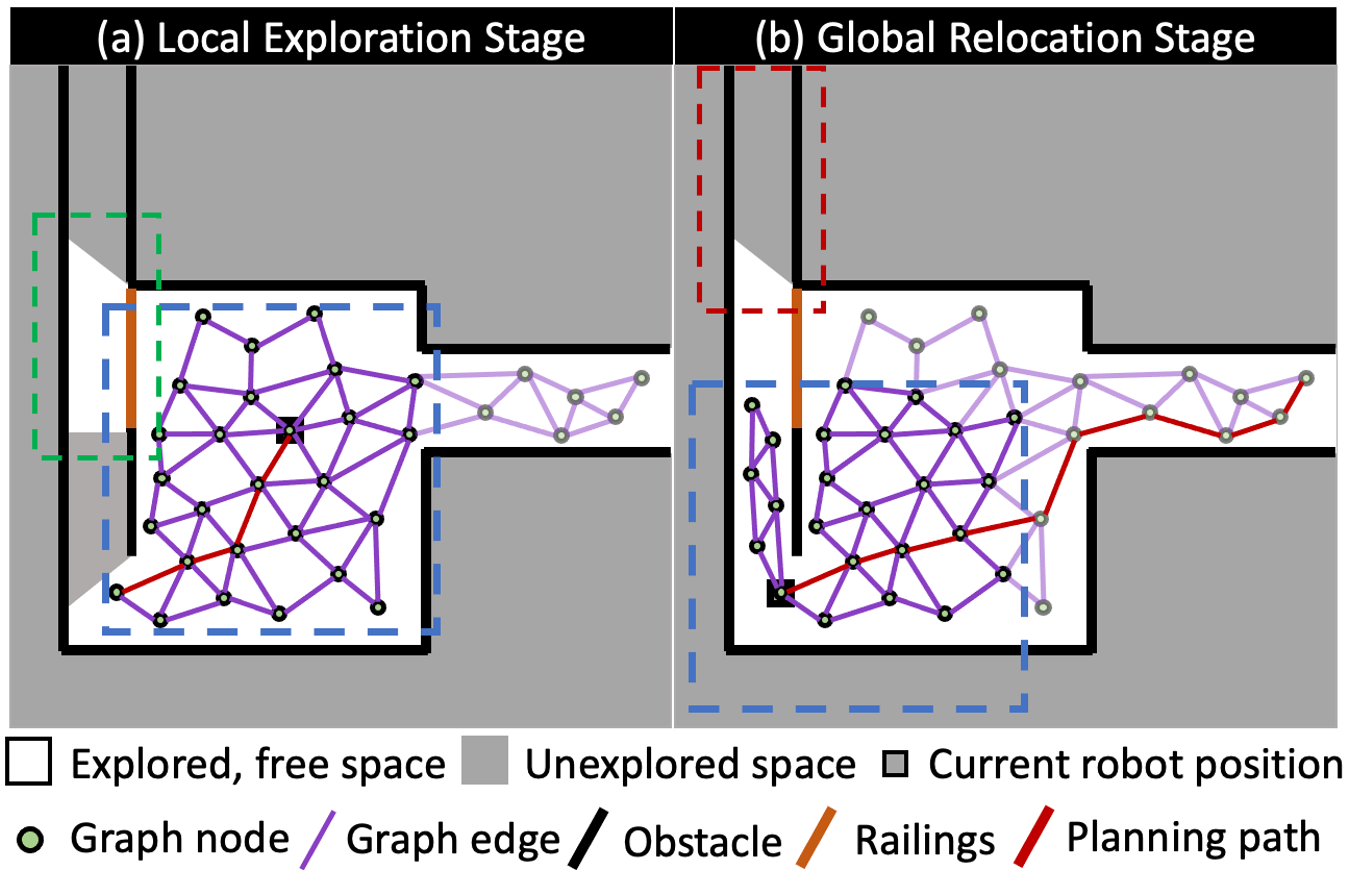}
		\caption{
		An example of incomplete exploration due to the failure of the local planner in the existing local exploration plus global relocation paradigm. (a) The area behind the railing (dotted green box) can be mapped but no more new nodes can be expanded by the local planner (due to the detected railing obstacle by the LiDAR sensor). At this moment, the robot planned a path (red curve) to the viewpoint with the highest exploration gain. 
		(b) After the robot bypasses the railing (stopping at the little dark-square location), it can no longer find any unknown area around the viewpoint in the local sliding window. Therefore, the robot considers this region as explored and goes to the next viewpoint with the highest exploration gain.  
		In this case, the unexplored area above (dotted red box) is permanently omitted. 
        }
		\label{railing}
	\end{figure}

\begin{figure}
        \centering
        \includegraphics[width=0.8\linewidth]{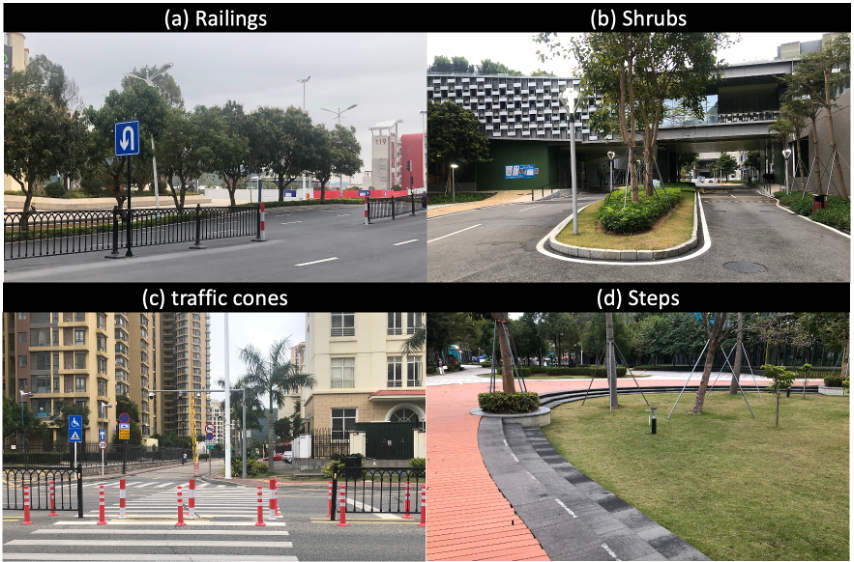}
		\caption{Some types of obstacles may cause incomplete exploration by the existing local exploration plus global relocation paradigm. }
		\label{shrubs}
	\end{figure}

    The above paradigm of local exploration plus global relocation \cite{dang2020graph, kulkarni2021autonomous, dsvp} has shown quite promising robotic exploration performance nowadays. However, the local planner can result in incomplete exploration in an unknown environment. For instance, Fig.\ref{railing} shows one failure example, where the area marked by dotted red lines is omitted by the local planner. This kind of incomplete exploration due to local-planner omission can easily occur in environments where obstacles can be detected, but do not completely block the LiDAR sensor with part of the areas behind the obstacles observable, such as penetrable obstacles (railings), low positive obstacles (shrubs, traffic cones), and negative obstacles (steps or ditches), etc.. Fig. \ref{shrubs} shows these scenarios, which are usually very common in outdoor environments. 
    
    
    
    
    
    
    Fig.\ref{railing} shows us a scenario that the existing paradigm of local exploration plus global relocation can fail in fully exploring an unknown environment, where although there is no unknown region around the nodes in the sliding window, some nodes (e.g., the ones just above the robot) still need to be visited. The underlying reason has not been revealed by the existing methods so far. To our best belief, it is the way of calculating the volumetric gain in the existing paradigm of local exploration plus global relocation that results in the failure of a complete environment exploration. We believe that it is inappropriate for the SOTA RRT-based exploration methods \cite{dang2020graph, kulkarni2021autonomous, dsvp} to calculate the volumetric gain as the volume of the unknown region around the viewpoint. It is well-known \cite{dsvp} that the exploration gain decides the next best exploration destination, and it is usually composed of the volumetric gain, distance, and exploration direction between the current robot location and the next potential viewpoint to be visited. 
    As one important component of the exploration gain, an inappropriate calculation of the volumetric gain inevitably results in an inaccurate exploration gain, which in turn arouses an incomplete exploration. 

    
    The volumetric gain of the existing methods is essentially calculated based on the sensor's observation. 
    In our opinion, the volumetric gain should be calculated according to the positional configuration of the viewpoints rather than the volume of the unknown region around the viewpoints. 
    For brevity, we call the existing way of calculating volumetric gain \textit{Unknown Gain}.

\begin{figure}
		\centering
		\includegraphics[width=0.9\linewidth]{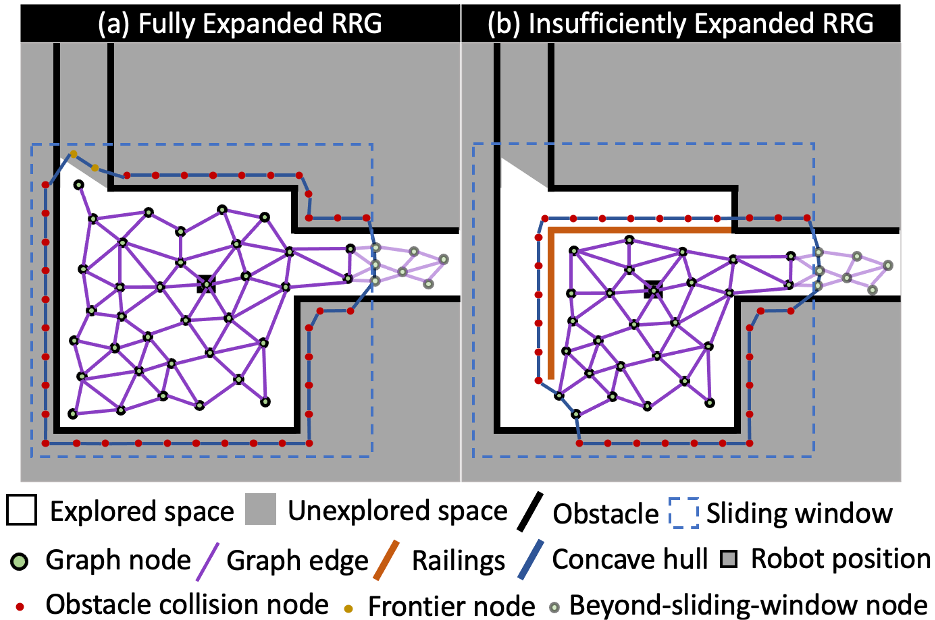}
		\caption{
		(a) A fully expended RRG can be considered as a concave hull wrapped by expansion-failure nodes. 
		RRG cannot expand outward any further until the robot explores new areas. 
		(b) An insufficiently expended RRG cannot be wrapped only by expansion-failure nodes to form a concave hull. 
		The nodes of the concave hull that belong to RRG are the locations where the omission occurs, meaning that we need to continue exploring around these nodes. 
		It can be proved that assuming that the RRG is wrapped by expansion-failure nodes but can still expand outward. 
		The RRT must pass through the space where the expansion-failure nodes are located when expanding, which is contradictory to the definition of expansion-failure. Therefore our inference is reasonable. 
        }
		\label{concave-hull}
	\end{figure}

    To address the above issues, we propose a concave-hull based method for volumetric gain calculation. 
    (1) We record the nodes where the RRT expansion fails. 
    The expansion-failure nodes can consist of three types, i.e., obstacle-collision nodes, frontier nodes, and beyond-sliding-window nodes.
    A concave hull can be constructed to enclose all RRG nodes and expansion-failure nodes. 
    We show that a fully expanded RRG can be wrapped by the three types of expansion-failure nodes (Fig. \ref{concave-hull}(a)). 
    In contrast, an insufficiently expanded RRG cannot be wrapped by these three types of nodes, and there will be a gap consisting of some RRG nodes on its concave hull, as shown in Fig. \ref{concave-hull}(b).
    When the gap width exceeds the diameter of the robot, it means that there is an omission of some unexplored areas which should have been investigated by the robot. 
    (2) We propose a new way of calculating the volumetric gain, called \textit{Graph Gain}. 
    We calculate the number of voxels around the viewpoint that are outside the concave hull instead of the number of unknown state voxels around the viewpoint.
    In this case, the volumetric gain is limited by the RRG's expansion range and no longer be determined by sensor observations. 
    Our exploration method is evaluated against the current RRT-based methods in a benchmark environment \cite{cmuexploration}.
    In the evaluated environment, our method significantly outperforms the state-of-the-art methods in terms of both running time and robustness. 
    
    The remainder of this paper is organized as follows.
    Section II outlines the related works.
    The details of our method are given in Section III.
    Experimental results are shown in Section IV, followed by conclusions in Section V.
    
    \section{Related Works}
    Existing exploration frameworks rely on either geometric frontiers or sampling frontiers, with which planners are required to plan collision-free paths towards the selected frontiers. 
    Methods based on geometric frontiers require searching in the map, while efforts have focused on reducing the search space.
    Yamauchi et al \cite{yamauchi1997frontier} proposed the seminal work to find the frontiers by searching the entire explored area during map update. 
    The Wavefront Frontier Detection \cite{keidar2012robot} reduces the search space from the entire explored space to the free space. 
    The Expanding-Wavefront Frontier Detection \cite{quin2014expanding} only incrementally searches for the latest explored areas. 
    Hess et al \cite{hess2016real} perform frontier detection in the submaps, reducing the search space to the total frontier length of all previous submaps and excluding the effect of graph optimization on the map. 
    Sun et al \cite{DBLP:conf/iros/SunW0SYK20} reduces the search space to the frontier length of the latest modified submaps. 
    These methods aim to detect all geometric frontiers, and they are mostly limited to 2D space and difficult to extend to 3D space. 
    
    With the popularity of UAVs and the need for large-scale environmental exploration, 
    the ``next-best-view" concept was proposed \cite{connolly1985determination}. 
    Oriolo \cite{oriolo2004srt} and Freda \cite{freda2005frontier} introduced a probabilistic planning method called Sensor-based Random Tree, a variant of RRT, to effectively sample the map. 
    The Multiple Rapidly-exploring Randomized Trees \cite{umari2017autonomous} enhances the local exploration efficiency by using global and local RRTs. 
    Receding Horizon Next-Best-View method (NBVP) \cite{bircher2016receding} and its variants \cite{dang2018visual, papachristos2017uncertainty} proposed online methods to find the optimal branch of random trees for exploration. 
    
    Accelerated by the DARPA Subterranean Challenge \cite{DARPA}, many robust and efficient underground exploration methods have been proposed. 
    The Graph-Based exploration path Planner (GBPlanner) \cite{dang2020graph} and Motion-primitive Based exploration path Planner (MBPlanner) \cite{dharmadhikari2020motion} effectively limits the search space of the RRT by using periodic sliding windows, allowing the robot to explore at a speed of 2m/s. 
    On this basis, the Dual-Stage Viewpoint Planner (DSVP) \cite{dsvp} retains the viewpoints in the overlapping area between adjacent sliding windows to reduce the cost for the RRT to re-expand. 
    Its variant \cite{ada} achieves incremental expansion by dynamically adjusting the sliding window size and sampling density, further reducing the expansion cost of RRT. 
    TARE planner \cite{tare} involves a hierarchical framework. 
    Smooth local trajectories are constructed by solving the traveling salesman problem (TSP) for random viewpoints, while only rough routes are retained for global trajectories to improve exploration efficiency. 
    GBPlanner2 \cite{kulkarni2021autonomous} added multi-robot collaboration and handling of steep slopes to the GBPlanner and won the challenge. 
    
    During the development of RRT-based exploration methods in 3D space, the calculation of the exploration gain has been improved. 
    NBVP proposes that the exploration gain $E_{gain}(v_{k})$ of each node $v_{k}$ in the RRT is the exploration gain of its parent node plus the volumetric gain $V_{gain}(v_{k})$ of itself with a scale factor $\lambda$ penalizing the distance cost between two nodes. 
    \begin{align}
        E_{gain}(v_{k})= E_{gain}(v_{k-1})+V_{gain}(v_{k})e^{-\lambda dis(v_{k-1},v_{k})}
    \end{align}
    GBPlanner and DSVP take into account the similarity of the current branch to the selected branch in the previous iteration, which reflects the direction of exploration.
    \begin{align}
        E_{gain}(v_{k})=e^{-\lambda_{1}sim(b_{v_{i}}, b_{pre})}\sum_{i=1}^{k} V_{gain}(v_{i})e^{-\lambda_{2} dis(v_{1},v_{i})}
    \end{align}
    where $sim(b_{v_{i}}, b_{pre})$ indicates the similarity between the branch where the current node $v_{i}$ is located and the branch selected in the previous iteration, and $\lambda_{1}$, $\lambda_{2}$ are the tuning factors. 
    According to this exploration gain, robots prefer the nodes just ahead of them, thereby reducing the costs associated with steering. 
    Although these methods consider a variety of factors that affect exploration behavior, they have little improvement in the calculation of the volumetric gain around the viewpoint. 
    Almost all methods calculate the volumetric gain as the number of unknown-state voxels around the viewpoint. 
    
    \section{Methods} 
    
    \begin{figure}
		\centering
		\includegraphics[width=1\linewidth]{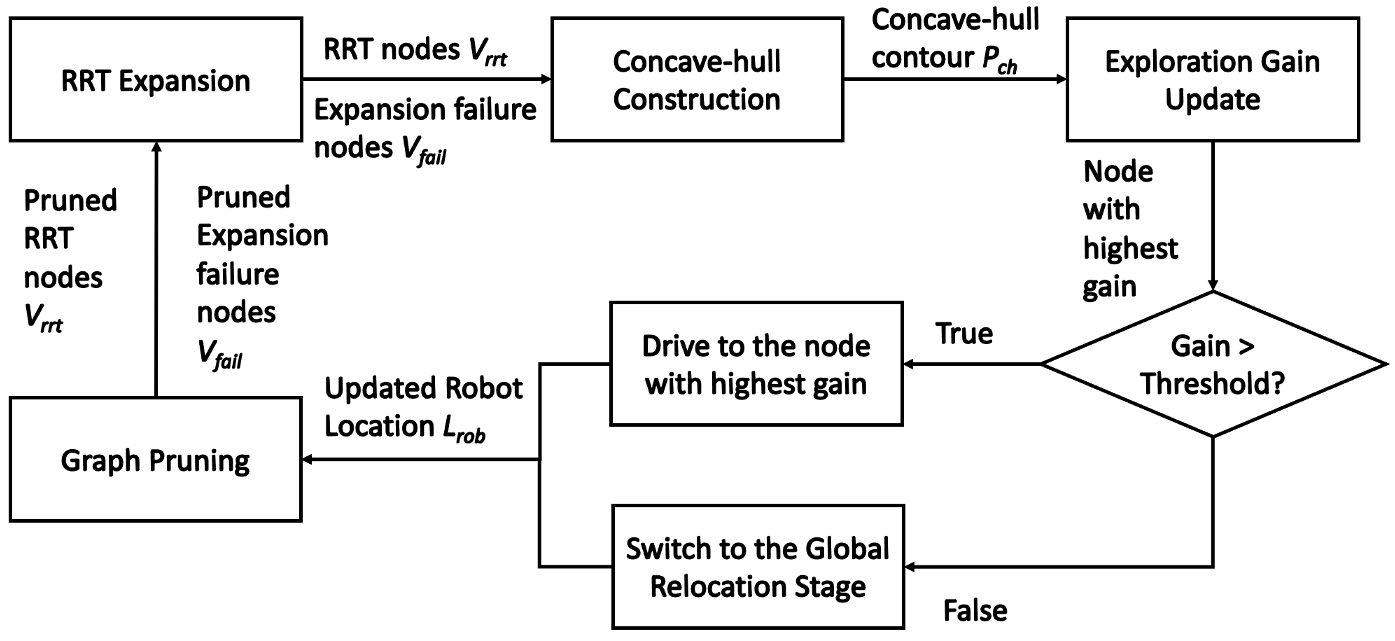}
		\caption{
		The system diagram of local exploration stage. 
        }
		\label{flow}
	\end{figure}
    
    Fig. \ref{flow} shows the main flow of the local exploration stage of our method, which is based on the backbone of the DSVP method \cite{dsvp}. 
    We record the nodes $V_{rrt}$ of the RRT and expansion-failure nodes $V_{fail}$ in the RRT expansion module, and combine $V_{rrt}$ and $V_{fail}$ together to form the set $V_{ch}$: $V_{ch} = V_{rrt} \cup V_{fail}$. 
    The nodes in $V_{ch}$ are classified into $3$ types as described above: ``successful", ``occupied", and ``unknown", and then down-sampled. 
    
\begin{figure}
		\centering
		\includegraphics[width=0.9\linewidth]{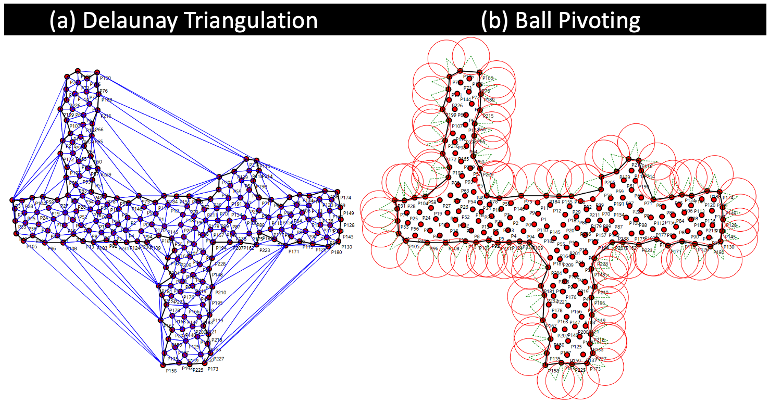}
		\caption{
		(a) Schematic diagram of concave hull construction based on Delaunay triangulation. 
		The boundaries of triangulation (blue) are the exterior edges, and the final concave hull (black) can be obtained by continuously removing the longest exterior edge whose length is greater than $R$. 
        (b) Schematic diagram shows the use of a ball of diameter $R$ rolled over the concave hull. 
		The distance between any two consecutive nodes on the concave hull is less than $R$, the minimum passing width of the robot that we set, meaning that we can determine whether the robot can pass between two nodes by querying the type of two consecutive nodes. 
		If we increase $R$, we cannot determine that the robot cannot pass even if two consecutive nodes are obstacle-collision nodes, while decreasing $R$ will lead to an increase in computational cost.
        }
		\label{delaunay}
	\end{figure}

    We consider that the concave hull of $V_{ch}$ represents the outermost layer of the RRT expanding outward and can be used to delineate the RRT expanded and unexpanded spaces. 
    So we construct a concave hull enclosing all the nodes in $V_{ch}$ based on the method introduced in \cite{duckham2008efficient} with a complexity of $O(NlogN)$, where $N$ is the number of nodes in $V_{ch}$. 
    Specifically, after performing Delaunay triangulation\cite{barber1996quickhull} on the set $V_{ch}$, the concave hull can be obtained by repeatedly removing the longest exterior edge from the triangulation until the lengths of all exterior edges are less than the maximum edge length $R$. 
    The exterior edges here are the edges belonging to only one triangle and not the common edges of two triangles, which is the boundary of the triangulation, as shown in Fig. \ref{delaunay}(a). 
    
    The most important parameter for computing the concave hull is the maximum edge length $R$ between two points on the edge. 
    We intuitively set $R$ to be twice of the robot size (to avoid collisions). 
    If two consecutive nodes on the concave hull are obstacle-collision nodes, it means that the robot cannot pass between the two nodes, as shown in Fig. \ref{delaunay}(b). 
    Then, we iterate over each node of the concave hull. 
    If there exist consecutive nodes of ``successful" or ``unknown" types, but their cumulative length is less than twice of the robot size, we consider their spacing too narrow for the robot to pass, while discarding these nodes. 
    In this way, we obtain the filtered concave hull polygon $P_{ch}$. 
    
    In the DSVP method, all voxels around the RRT node $L_{node}$ that are eligible for calculating the exploration gain are checked for their status (occupied, free or unknown) in the grid map. In general, these eligible voxels are within a circle that centers at $L_{node}$ with a certain radius. 
    
\begin{figure}
		\centering
		\includegraphics[width=0.8\linewidth]{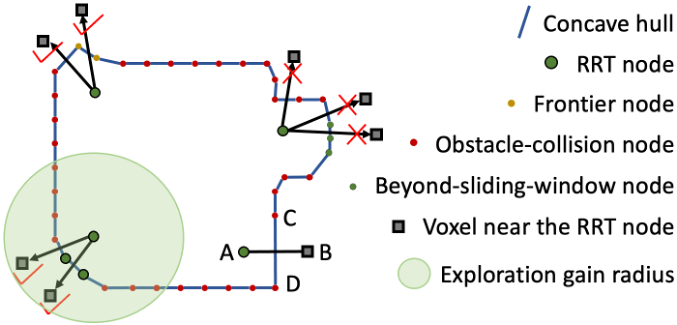}
		\caption{
		    An example illustrates the effect of different states of the nodes on the volumetric gain calculation.  
		    We find the edge of the concave hull that intersects the line linking the RRT node to the voxel. 
		    When both nodes on the edge are obstacle-collision nodes or beyond-sliding-window nodes, the voxel is not calculated as a volumetric gain. 
		    If the state of at least one node on the edge is ``successful" or ``unknown", the voxel is calculated as a volumetric gain. 
        }
		\label{graph-gain}
	\end{figure}

    For those voxels whose states are unknown, it is then checked whether there are obstacles between them and $L_{node}$. 
    If not, the number of such eligible voxels are used as the volumetric gain of the node $L_{node}$. 
    In contrast, in our method, all voxels that are eligible for calculating the exploration gain are checked if they are inside the concave hull. 
    For each of such eligible voxels that are outside the concave hull, we find the edge of concave hull where the concave hull intersects the line linking the voxel to the node $L_{node}$ (Fig.\ref{graph-gain}).
    If the two nodes of this specific types of edges are ``successful" or ``unknown" nodes, the number of such voxels are used as volumetric gain of the node $L_{node}$, as shown in Algorithm \ref{algorithm3}. 
    Fig.\ref{graph-gain} shows the effect of different states of the nodes on the volumetric gain calculation. 
    Since the RRT of the ground robot mostly extends in 2D, we only need to calculate the volumetric gain of the voxel projection on the ground plane.
    
    \begin{algorithm}
		\caption{Update Volumetric Gain}
		\label{algorithm3}
		\KwIn{ \\
		    \quad Concave Hull Polygon $P_{ch}$  \\ 
		    \quad RRT nodes $V_{rrt}$ \\
		    \quad robot location: $L_{rob}$ \\
        }
		\KwOut{ \\
			\quad volumetric gain: $Gain_{v}$ \\
		}
		\SetKwIF{If}{ElseIf}{Else}{if}{}{else if}{else}{end if}%
		$Gain_{v}$ = 0 \\
		\For{ $L_{node}$ iterates in $V_{rrt}$}{
			\For{ $V$ iterates over all voxels within the allowed range of $L_{node}$}{
			    \If{pointInPolygon($V$, $P_{ch}$) = FALSE and checkIntersection($L_{node}$, $V$) = True}{$Gain_{v}$ += 1}
			}
		}
	\end{algorithm}
    

    We use the PNPOLY method \cite {haines1994point} to determine whether a point is inside a concave hull with a complexity of $O(N)$, where $N$ is the number of nodes in the polygon $P_{ch}$. 
    We use the cross product to find the edge of the concave hull that intersects the line linking the node $L_{node}$ to an eligible voxel $V$. 
    As shown in Fig. \ref{graph-gain}, for the line segments AB and CD, if $\Vec{AC} \times \Vec{CD}$ and $\Vec{BC} \times \Vec{CD}$ are different signs, it shows that A and B are on both sides of the line CD (but do not mean that line segment AB intersects CD because line segment CD is of finite length). 
    If $\Vec{CB} \times \Vec{AB}$ and $\Vec{DB} \times \Vec{AB}$ are also different signs, then it means that the two line segments intersect, as shown in Algorithm. \ref{algorithm4} line 4-5. 
    
    
    \begin{algorithm}
		\caption{Check Intersection}
		\label{algorithm4}
		\KwIn{ \\
		    \quad Concave Hull Polygon $P_{ch}$  \\ 
		    \quad Viewpoint $L_{node}$ \\
		    \quad Voxel $V$ \\
        }

		\SetKwIF{If}{ElseIf}{Else}{if}{}{else if}{else}{end if}%
        Line1 = ($V$, $L_{node}$) \\
		\For{ $P_{i}$, $P_{i-1}$ iterate the two endpoints of each edge of the Polygon $P_{ch}$}{
		    Line2 = ($P_{i}$, $P_{i-1}$) \\
			 \If{(($V$-$P_{i}$)$\times$($P_{i-1}$-$P_{i}$))*
			      ($L_{node}$-$P_{i}$)$\times$($P_{i-1}$-$P_{i}$))$<0$ \\
			 and (($P_{i-1}$-$V$)$\times$($L_{node}$-$V$))*
			      ($P_{i}$-$V$)$\times$($L_{node}$-$V$))$<0$}{
			      \If{($P_{i}$.label = SUCCESSFUL or UNKNOWN \\ and \\
			         $P_{i-1}$.label = SUCCESSFUL or UNKNOWN}{
			         return True \\
			      }
			      \Else{
			          return False \\
			      }
			 }
		}
	\end{algorithm}
    
    When the sliding window is updated, we prune $V_{ch}$. The pruned range is slightly larger than the updated sliding window range.
    Nodes outside the updated sliding window are marked as ``beyond-sliding-window", while ``successful" and ``occupied" nodes within the window are left unchanged, and ``unknown" nodes are ignored. 
    In this way, the volumetric gain can be correctly calculated when the robot passes again over the area that has already been explored.

    \section{Experimental Results}
    
\begin{figure}[ht]
		\centering
		\includegraphics[width=0.8\linewidth]{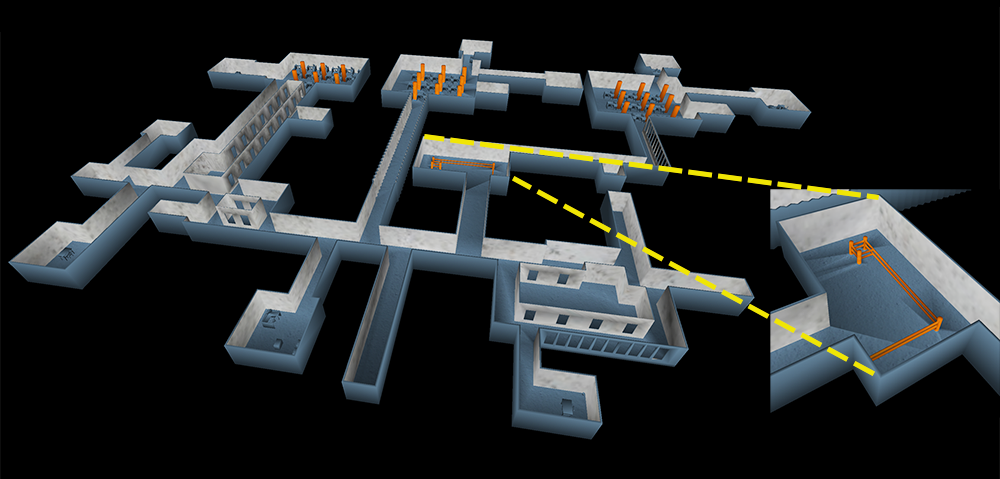}
		\caption{
		Indoor Corridors Environment (130m x 100m) \cite{cmuexploration}. 
		A close-up of the room with railings is presented in the lower right corner. 
        }
		\label{indoor}
	\end{figure}

    We compare the volumetric gains obtained by our method (Graph Gain) and the DSVP method (Unknown Gain) at different locations in the benchmark simulation environment \cite{cmuexploration}, as shown in Fig. \ref{indoor}. 
    
\begin{figure}[ht]
		\centering
		\includegraphics[width=1\linewidth]{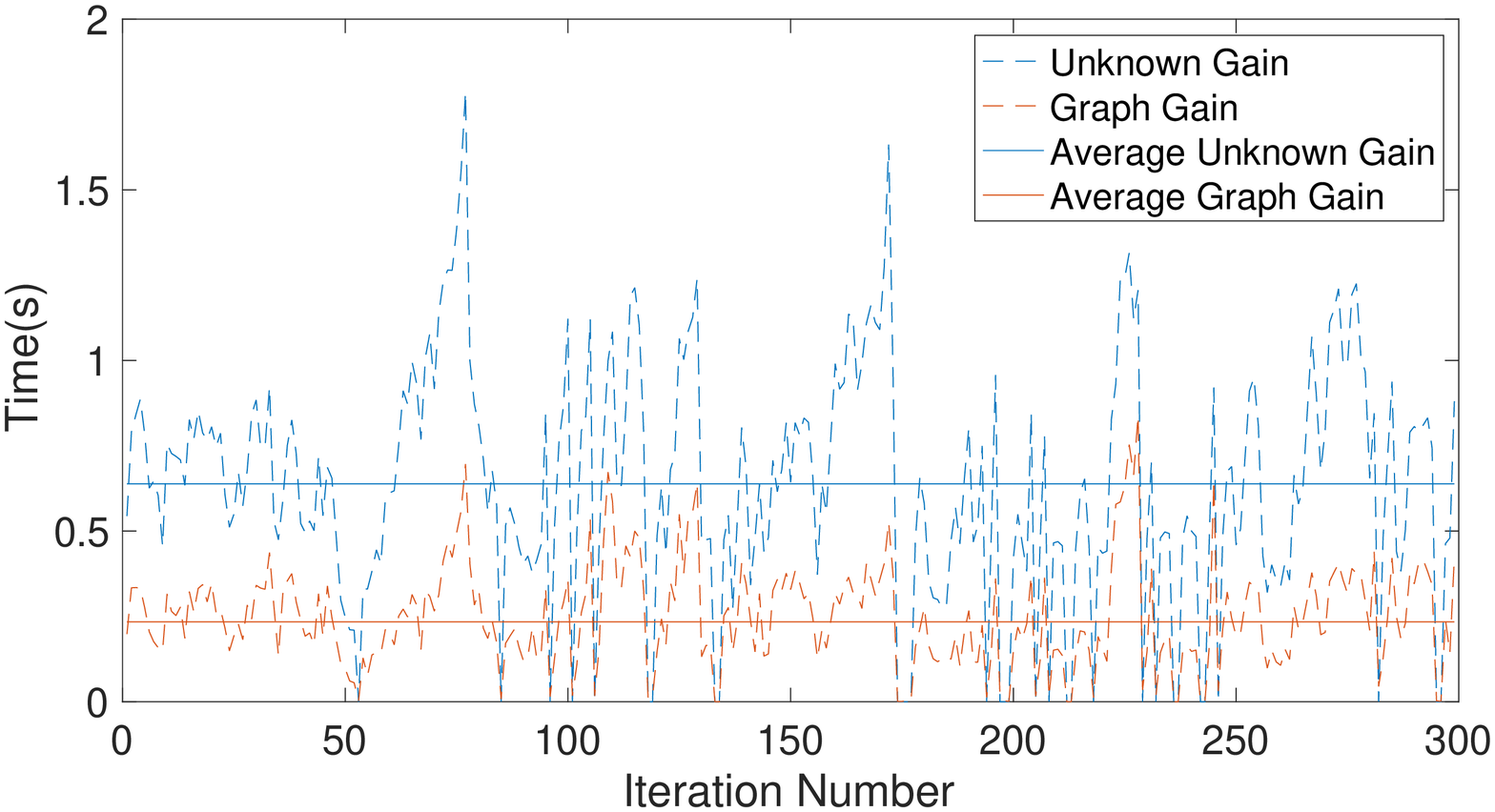}
		\caption{
		    Comparison of running time per iteration and average running time for Unknown Gain and Graph Gain. 
        }
		\label{time-compare}
	\end{figure}

    We compare the running time of Unknown Gain and Graph Gain. 
    As shown in Fig. \ref{time-compare}, we show the running time per iteration and average running time of Unknown Gain and Graph Gain during one of the explorations processes in the benchmark simulation environment, where the Graph Gain includes the concave hull construction module and the volumetric gain update module. 
    It shows that the average running time of Graph Gain is about $38.8\%$ of Unknown Gain. 
    We also calculated the average running time of both methods over 10 explorations. 
    Our Graph Gain method runs in $38.4\%$ of the running time of the Unknown Gain method. 
    
	\begin{figure*}[htbp]
		\subfigure[3D grid map at location 1]{
			\includegraphics[width=0.31\linewidth]{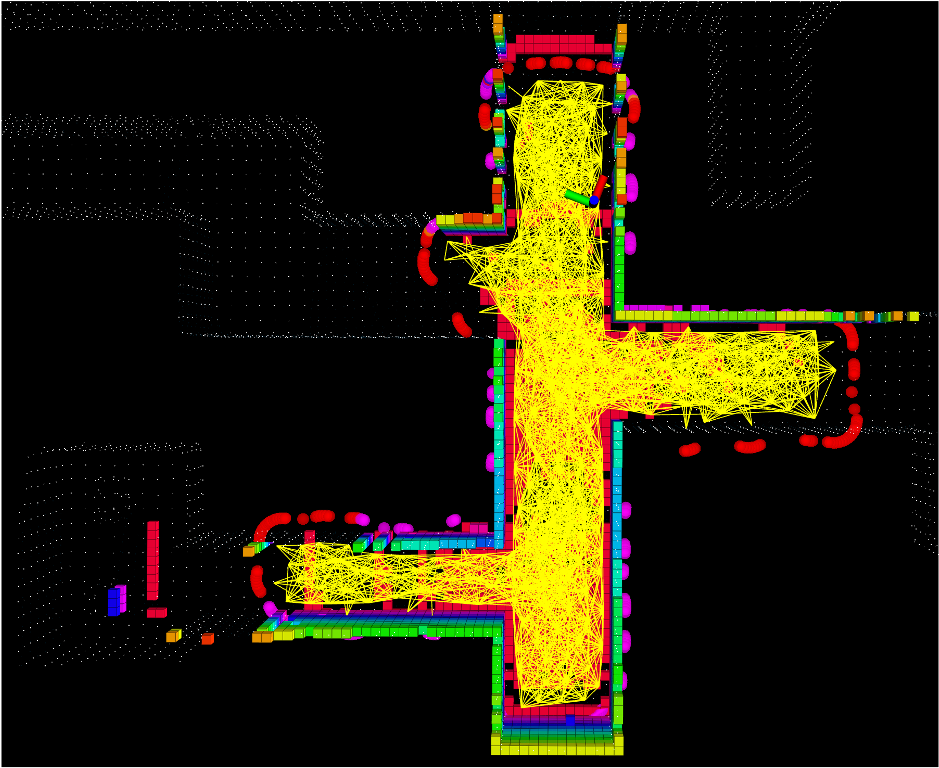}
		}
		\subfigure[3D grid map at location 2]{
			\includegraphics[width=0.31\linewidth]{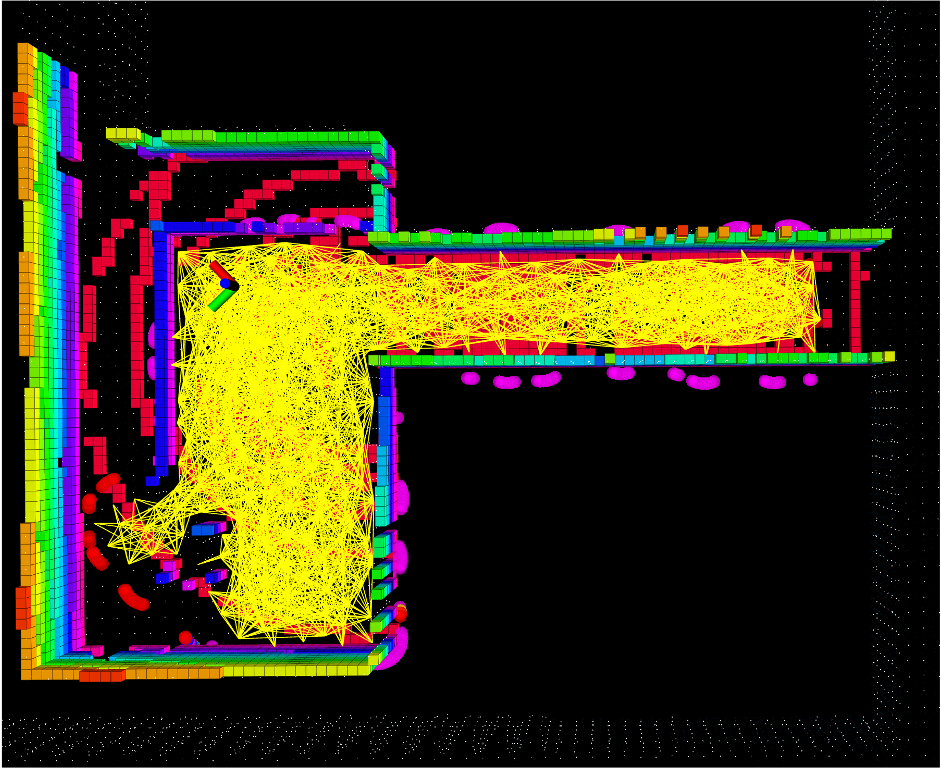}
		}
		\subfigure[3D grid map at location 3]{
			\includegraphics[width=0.31\linewidth]{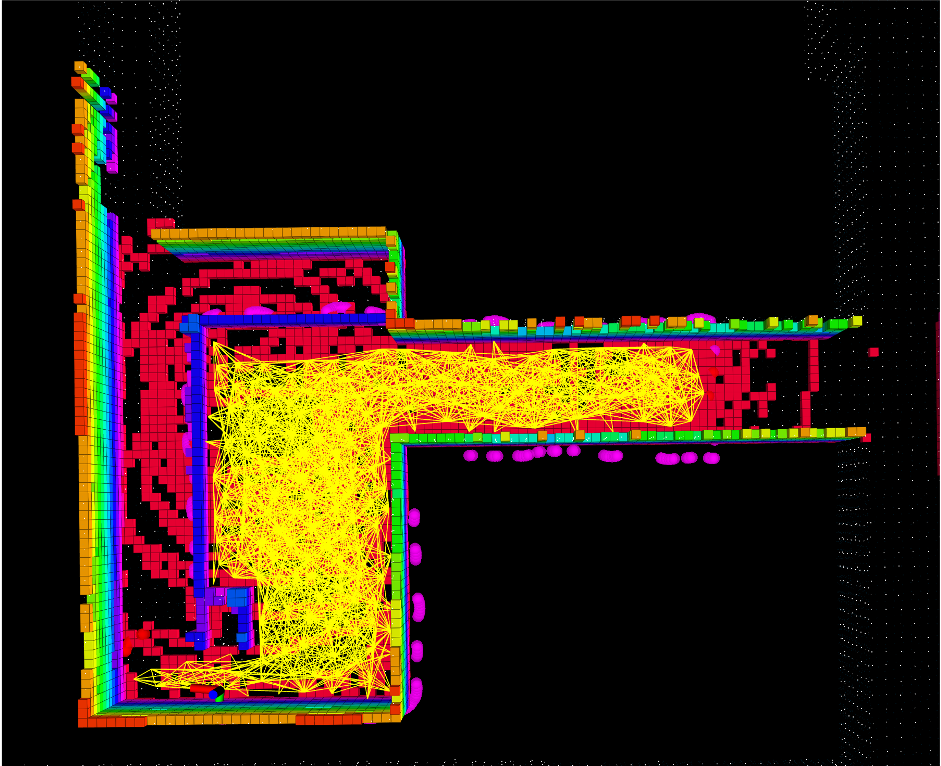}
		}
		\subfigure[Concave hull at location 1]{
			\includegraphics[width=0.31\linewidth]{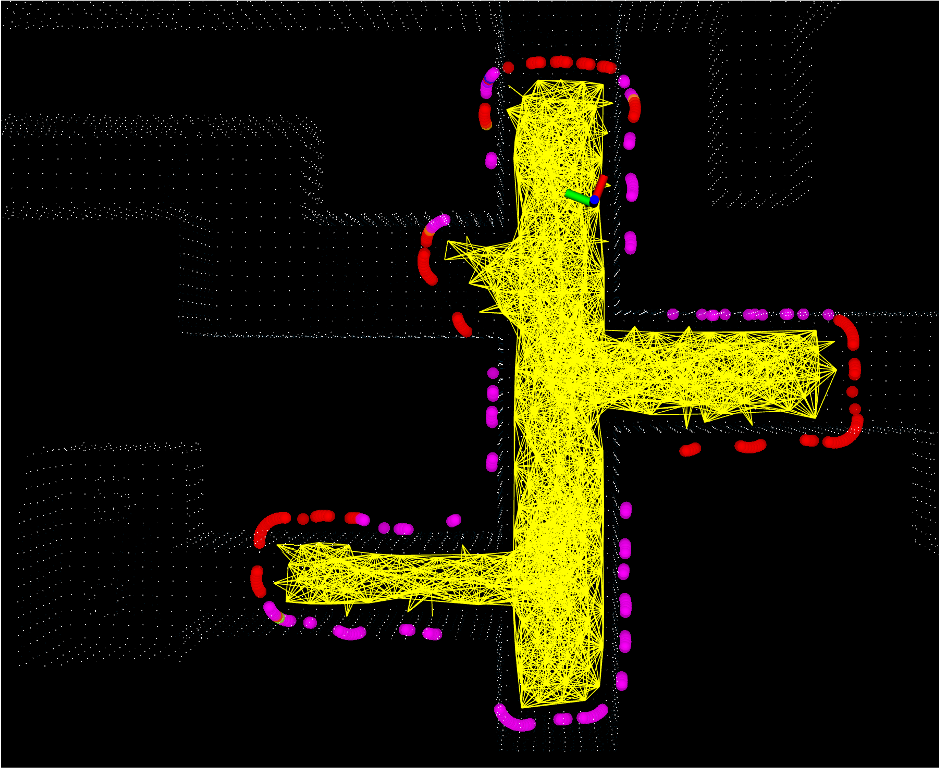}
		}
		\subfigure[Concave hull at location 2]{
			\includegraphics[width=0.31\linewidth]{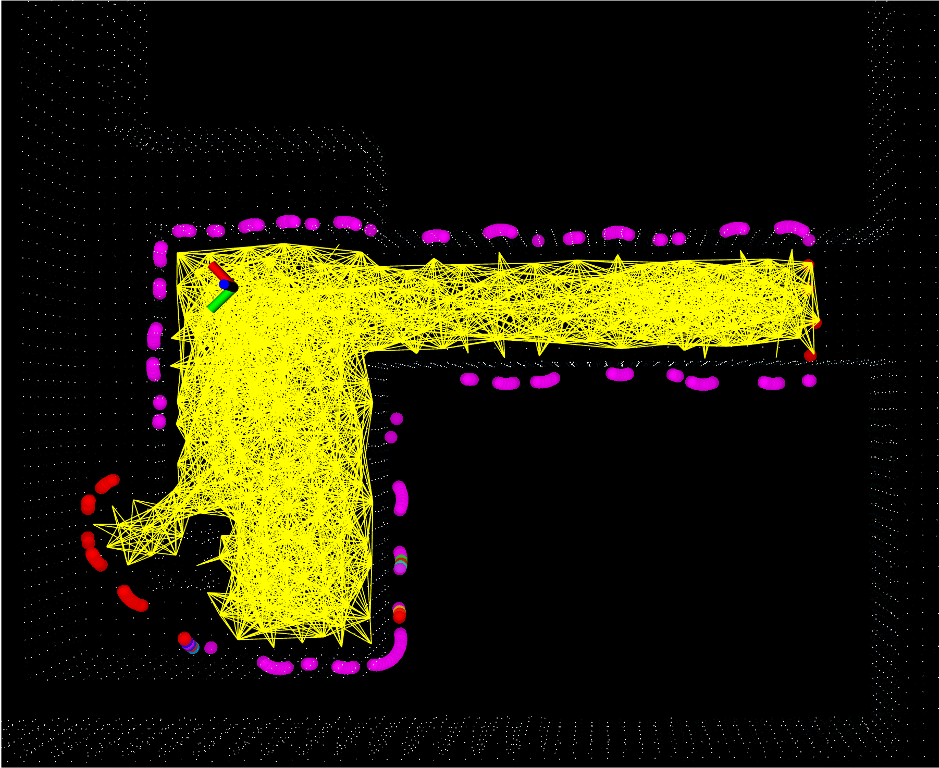}
		}
		\subfigure[Concave hull at location 3]{
			\includegraphics[width=0.31\linewidth]{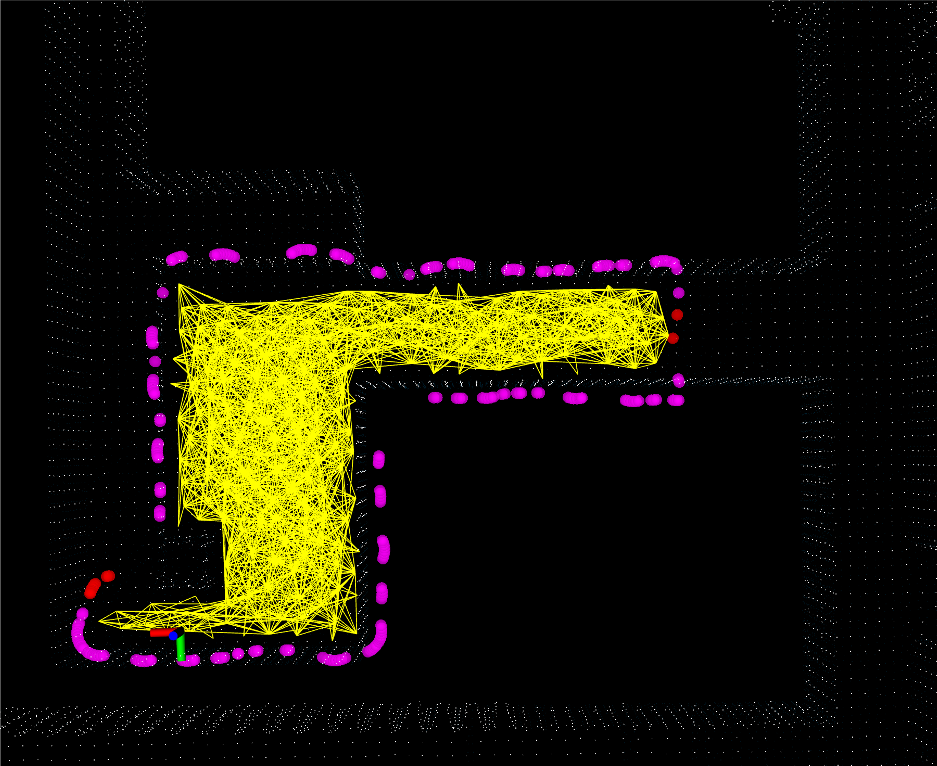}
		}
		\subfigure[Comparison of volumetric gain at location 1]{
			\includegraphics[width=0.31\linewidth]{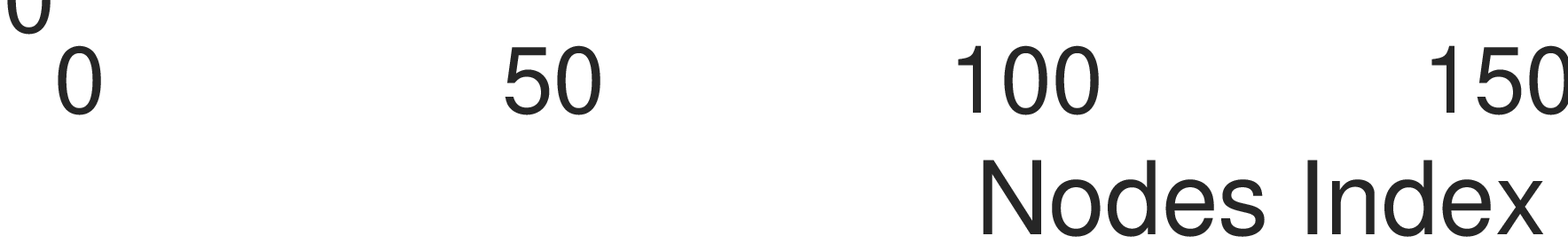}
		}
		\subfigure[Comparison of volumetric gain at location 2]{
			\includegraphics[width=0.31\linewidth]{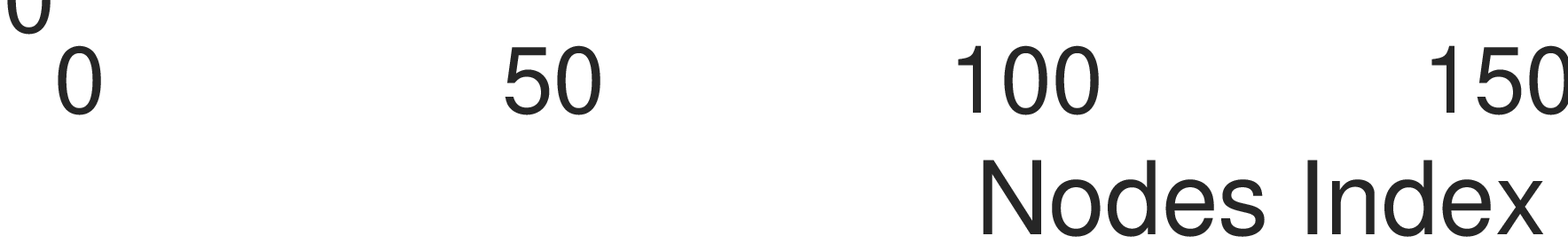}
		}
		\subfigure[Comparison of volumetric gain at location 3]{
			\includegraphics[width=0.31\linewidth]{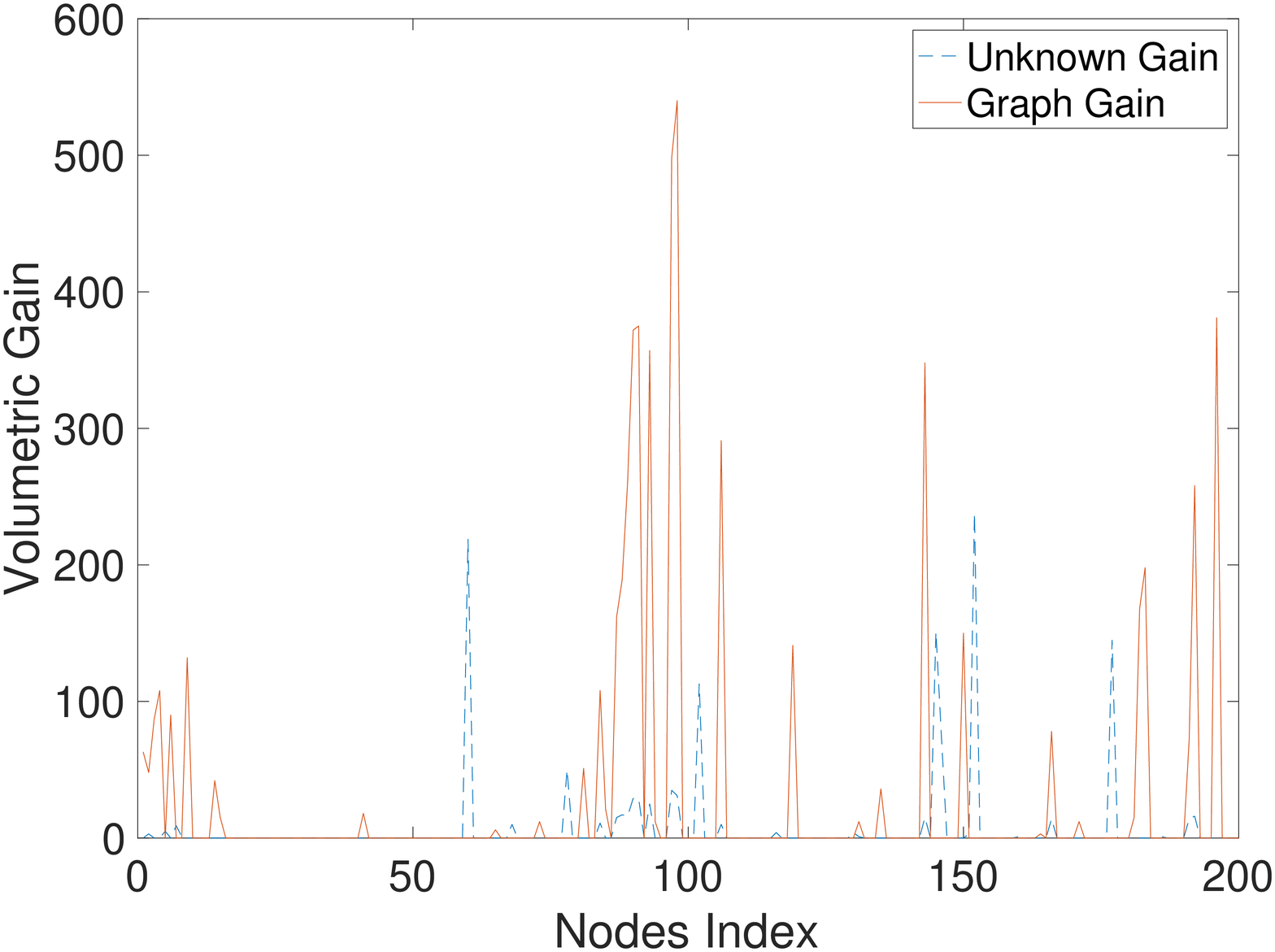}
		}
		\subfigure[Visualization of volumetric gain at location 1]{
			\includegraphics[width=0.31\linewidth]{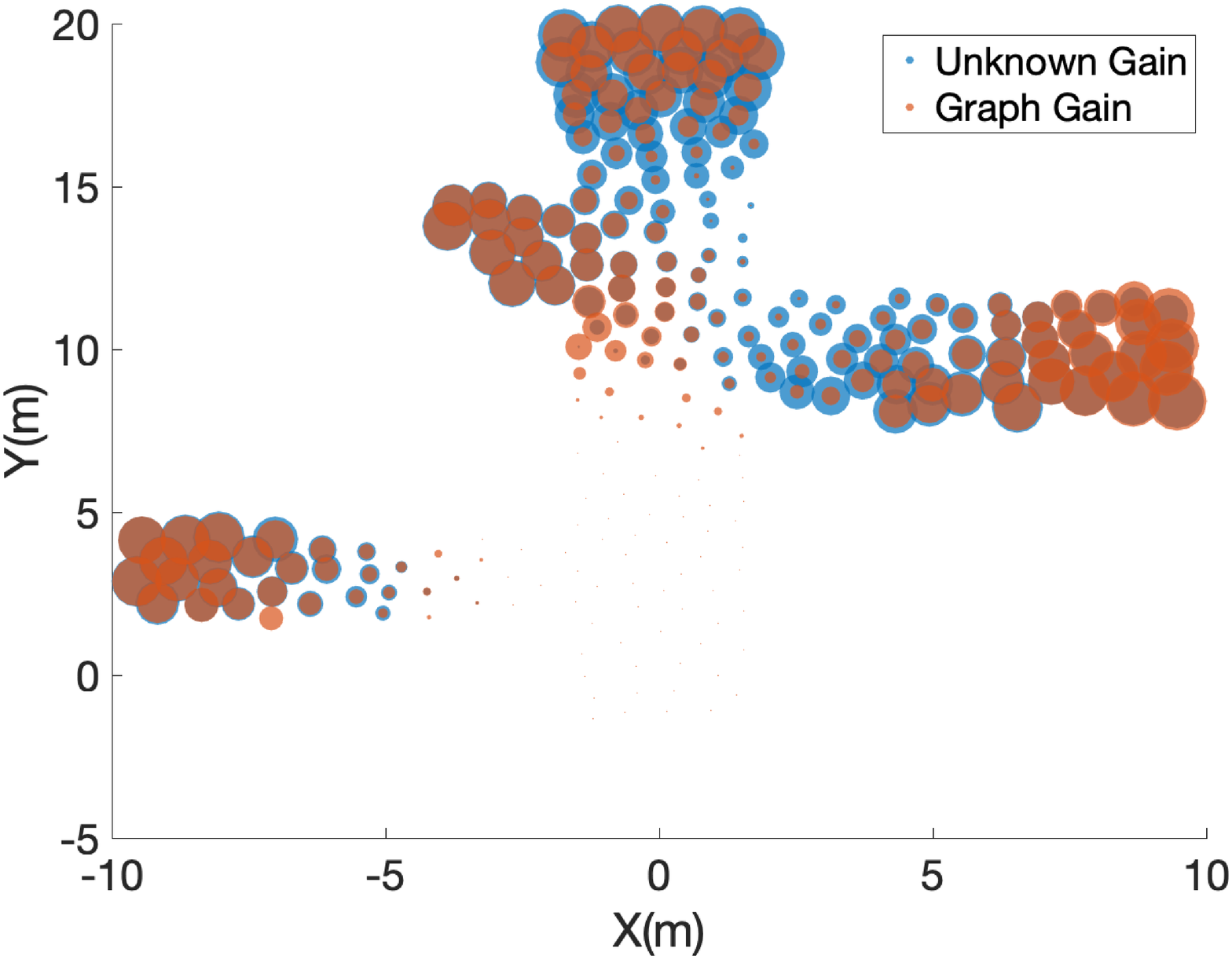}
		}
		\subfigure[Visualization of volumetric gain at location 2]{
			\includegraphics[width=0.31\linewidth]{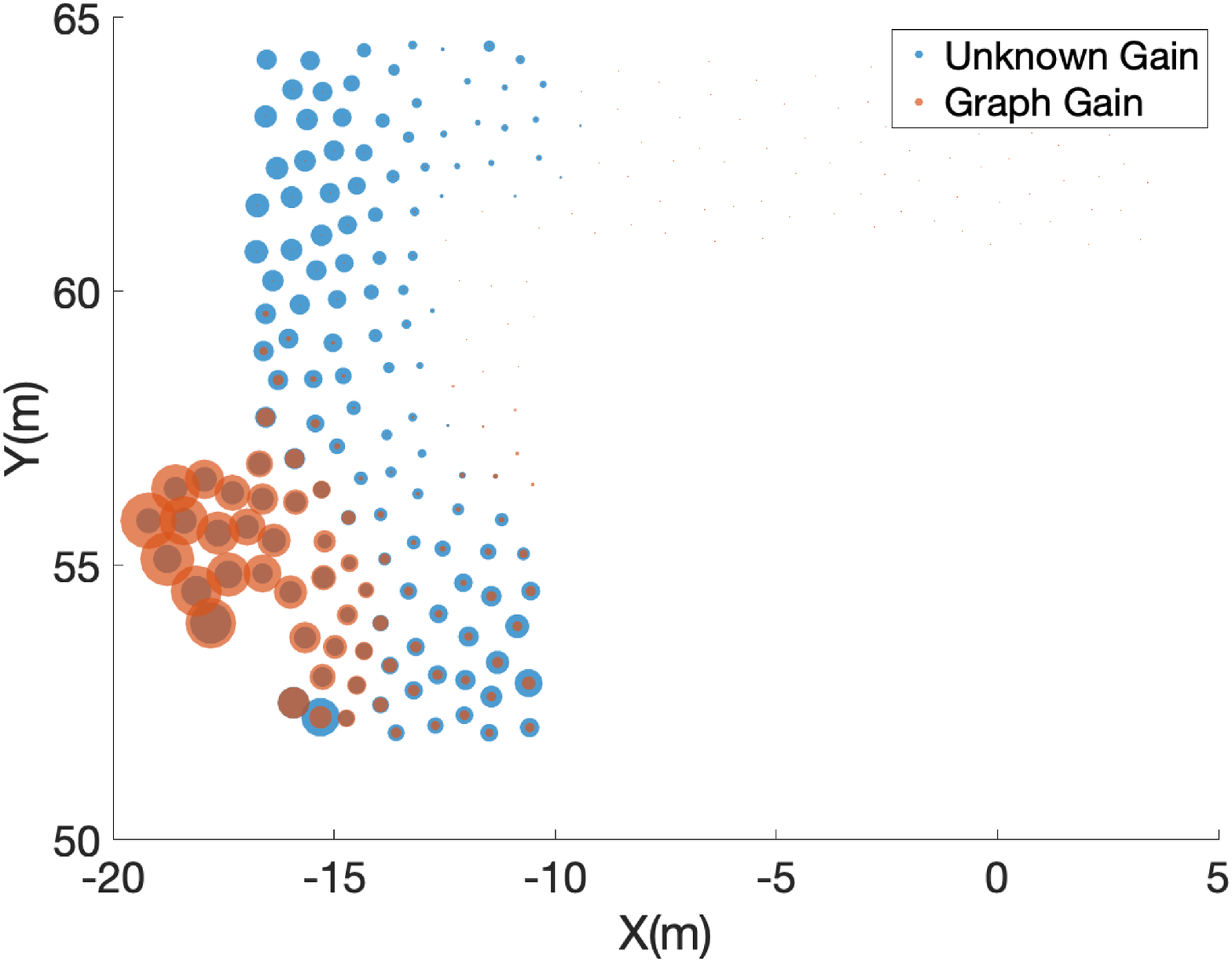}
		}
		\subfigure[Visualization of volumetric gain at location 3]{
			\includegraphics[width=0.31\linewidth]{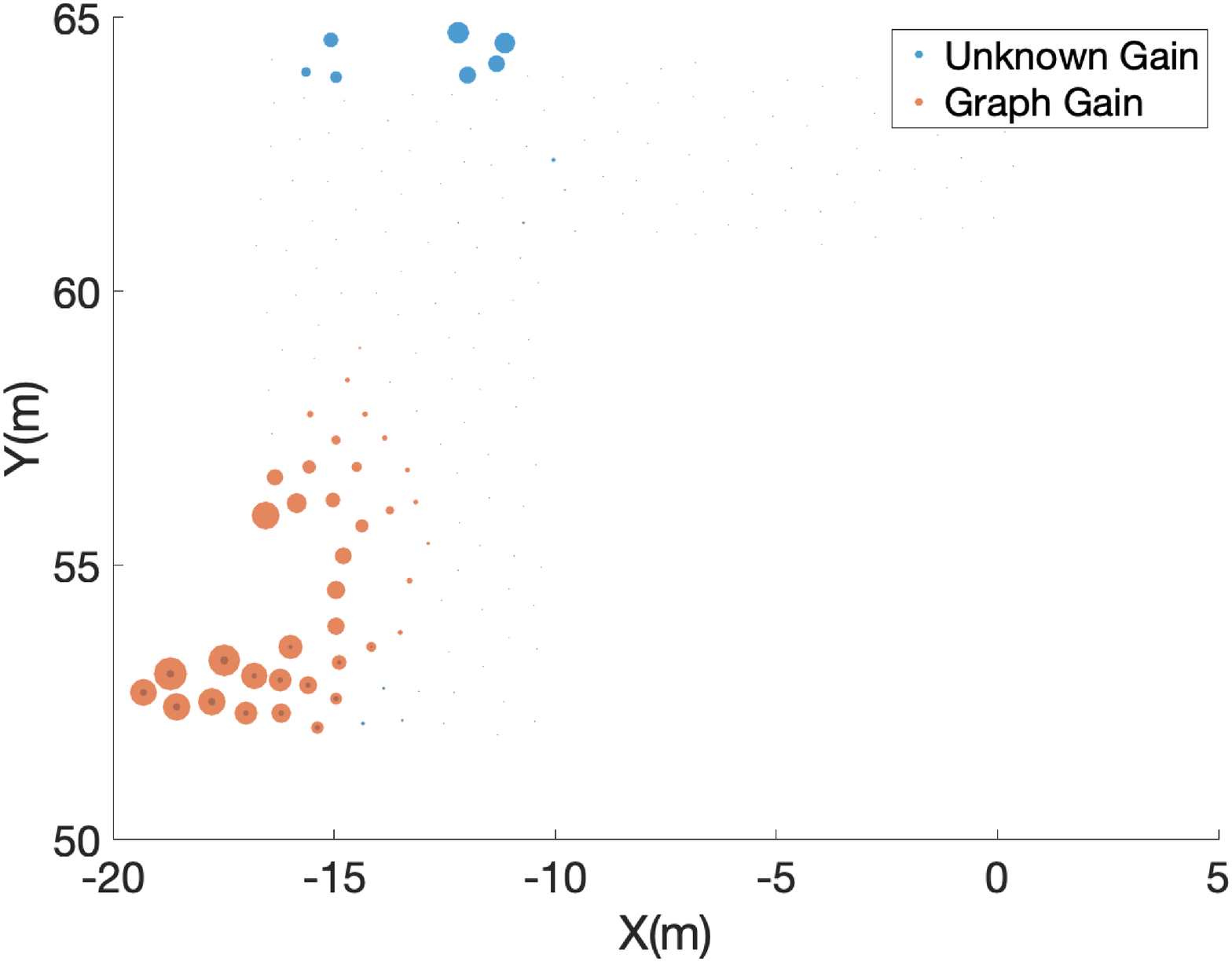}
		}
		\caption{The three columns compare the difference between Unknown Gain and Graph Gain at the three locations. 
        }
		\label{data}
	\end{figure*}
    
    In Fig. \ref{data}, the three columns show the differences between Graph Gain and Unknown Gain calculation of volumetric gain at different locations. 
    The common parameters are the same for both methods, which are the default parameters open-sourced by DSVP. 
    The first row shows the 3D grid maps created by the sensors at three locations. 
    The yellow graph is the RRG generated in the current sliding window. 
    The second row shows the constructed concave hull enclosing RRG and expansion-failure nodes, where the red nodes are frontier or RRG nodes and the purple nodes are obstacle-collision or beyond-sliding-window nodes. 
    The third row shows the volumetric gain of each node in the RRG calculated by Graph Gain and Unknown Gain, respectively. 
    The fourth row visualizes the volumetric gain of all nodes at their positions.
    For each circle, the center coordinates represent the coordinates of the node in the environment, and the radius represents the volumetric gain of the node. 
    (g) and (j) show that in the general corridor scene, the volumetric gains calculated by Graph Gain and Unknown Gain are similar. 
    (h) and (k) show that the RRG is blocked by the railing as it expands, so the node in front of the railing should no longer have volumetric gain to attract robots to explore there even though there are unknown regions behind the railing. 
    This results in little variation in volumetric gain between nodes using Unknown Gain, and it is difficult for the robot to find the next optimal exploration goal. 
    However, using Graph Gain to calculate the volumetric gain can effectively discern the next optimal exploration goal. 
    (f) and (i) show that when the robot bypasses the railing, the space behind the railing has already been mapped but not yet expanded by the RRT. 
    Thus Unknown Gain gives the wrong next exploration goal, which can be effectively avoided by Graph Gain. 
    
    \section{Conclusion}
    
    In this paper, we propose a method for calculating volumetric gain based on concave hulls. 
    We obtain the representation of the environment model of the RRT extended region by classifying the RRT nodes and the expansion-failure nodes. 
    Known and unknown spaces are divided by the inside and outside of the concave hull instead of the grid states of the 3D occupancy map. 
    Whether a voxel is calculated as a volumetric gain is determined by the state of the nodes at the ends of the edges where the concave hull intersects the line linking the voxel to the viewpoint, rather than detecting obstacles between the voxel and the voxel in the 3D grid map.
    The calculation of our volumetric gain no longer depends on the occupancy map constructed by the sensor but on the viewpoints configuration sampled by RRT, which avoids inefficient or even erroneous exploration behavior due to the inaccuracy of current volumetric gain calculation methods. 
    We validate the effectiveness of our method in a benchmark simulation environment and compared the average running time of our method to about $38.4\%$ of the current state-of-the-art method.

    \bibliographystyle{unsrt}
    \bibliography{bibfile}
	
\end{document}